# Predicting Early-Onset Colorectal Cancer with Large Language Models

Wilson Lau, PhD[1*], Youngwon Kim, PhD[1*], Sravanthi Parasa, MD[2], Md Enamul Haque, PhD[1], Anand Oka, PhD[1], Jay Nanduri[1]

[1] Truveta, Bellevue, WA; [2]Swedish Medical Center, Seattle, WA[1]

Abstract

*The incidence rate of early-onset colorectal cancer (EoCRC, age < 45) has increased every year, but this population is younger than the recommended age established by national guidelines for cancer screening. In this paper, we applied 10 different machine learning models to predict EoCRC, and compared their performance with advanced large language models (LLM), using patient conditions, lab results, and observations within 6 months of patient journey prior to the CRC diagnoses. We retrospectively identified 1,953 CRC patients from multiple health systems across the United States. The results demonstrated that the fine-tuned LLM achieved an average of 73% sensitivity and 91% specificity.*

Introduction

Colorectal cancer (CRC) is a significant public health concern, ranking as the second leading cause of cancer-related deaths and the 4th most common new cancer diagnosis in the U.S. in 2024.[1] While CRC has historically been considered a disease of older adults, there has been an increase in colorectal cancer diagnosed in individuals under 50. Between 2011 and 2019, CRC incidence rates increased by 1.9% per year in people younger than 50 years.[2,3] Furthermore, between 2012 and 2021, among individuals aged 20 to 49, the incidence of advanced-stage colorectal cancer increased by approximately 3% per year.[2,4] Because current screening guidelines recommend initiating routine CRC screening at age 45, younger adults are often diagnosed at later, less treatable stages.[5,6] Given these concerning trends, early identification of high-risk individuals is crucial for timely intervention and improved patient outcomes. Beyond standard screening, additional strategies are necessary to enhance the prevention and early detection of CRC in younger population. One promising approach is leveraging electronic health records (EHR) to identify individuals at increased risk.[7,8,9] By integrating diverse clinical and demographic data, EHR-based predictive approaches have the potential to enhance early detection efforts, increase awareness among both patients and healthcare providers, and facilitate timely medical evaluations, ultimately reducing the burden of EoCRC.

Colorectal Cancer Risk Prediction: Machine learning (ML) methods offer considerable advantages by modeling complex, non-linear relationships and interactions within large datasets. ML-based approaches in CRC prediction often demonstrate strong predictive performance and can handle noisy data, large sample sizes, and a higher number of predictive variables.[10,11,12] Despite these advantages, many ML-based models are still constrained by the incorporated features and the populations they are trained on.

Most existing ML studies on CRC focus on a limited set of well-established risk indicators, including demographic data (age, race, family history), lifestyle factors (BMI, alcohol consumption, red meat intake), and overt symptoms (rectal bleeding, blood in the stool, changes in bowel habits, abdominal pain, and blood count abnormalities).[11,13,14] While these factors are important, their utility for early detection is limited. Demographic-based models may not generalize well across diverse populations, and symptom-based models depend on clinical signs that typically emerge in later disease stages.[15] Younger individuals, in particular, may under-report or overlook symptoms, further delaying diagnosis.[16] This narrow focus potentially obscures a broader picture that could be revealed by considering a more comprehensive range of health record data. Additionally, the low prevalence of CRC in the general population presents challenges for ML models, making it difficult to achieve high predictive performance and develop robust risk assessment tools for real-world settings.

---

[*] equal contribution

A further limitation of existing ML studies for CRC is their reliance on a relatively small number of predictive variables, often fewer than 60 features.[10,12,15] This limitation may reduce their effectiveness for early detection, as they might fail to capture the full range of subtle risk factors. Moreover, the majority of CRC risk models were developed for individuals over 50, making them less applicable to younger adults. Among studies focusing on younger populations, the work by Zhen et al.[10] is most similar to our approach. They developed eight ML models to predict EoCRC in individuals aged 18–49, achieving a sensitivity of 84% and specificity of 73% using a random forest model. However, their study was conducted on a Chinese population from a single medical center, with a relatively small training cohort of 199 CRC patients. Additionally, their dataset contained a higher proportion of CRC patients (approximately 5%) than typically seen in real-world clinical settings, potentially limiting the generalizability of their findings.

Advancing CRC Prediction with Large Language Models: The recent advancement of large language models propels new predictive capabilities in clinical research.[17] LLMs have demonstrated promising performance in predicting various health conditions, including depression, sleep disorders, and stress levels[18] as well as in rare disease diagnosis,[19] showcasing their ability to uncover hidden patterns in clinical data. Unlike ML models, LLMs can process and analyze vast amounts of unstructured and structured EHR data, potentially identifying risk factors that may be overlooked in conventional feature-engineering approaches. One advantage of LLMs is their flexibility in generalizing across diverse populations and diseases, reducing dependence on specific sample sizes or populations—especially in base models without fine-tuning. Our study builds on these recent developments by being, to our knowledge, the first to investigate and employ LLMs for predicting EoCRC in individuals aged 18–44 across multiple U.S. health systems. By integrating LLMs into predictive modeling, we aim to enhance early detection efforts and contribute to more personalized risk assessment for younger individuals at risk of CRC.

Methods

The aim of this study is to predict the onset of CRC in the younger population between 18 and 44 years of age, by employing both statistical machine learning models (ML) and large language models (LLM). The inclusion criteria for the CRC patient group were: 1) confirmed diagnoses of CRC in the EHR; 2) diagnosed at age 18-44. The inclusion criteria for the Non-CRC control group were: 1) absence of CRC diagnoses; 2) age between 18 and 44. The exclusion criteria for both groups consisted of: 1) any personal or family history of CRC; 2) diagnoses of Crohns disease, Lynch syndrome, or ulcerative colitis. To train the machine learning models, we used patient conditions, lab results, and observations from the 2nd to the 7th month prior to the CRC diagnoses. We intentionally excluded patient data within the last month prior to diagnoses in order to avoid signals that were highly indicative of an impending diagnosis of CRC. As such, any patients without 7 months of medical conditions, lab results and observations would be excluded from the study. A set of ICD and SNOMED-CT codes related to each of the criteria were curated and reviewed by a gastroenterologist. We used these codes to identify the cohorts in the Truveta database where all patient data were already deidentified.

Dataset: A total of 1953 patients were identified for the CRC group, which was less than 1% of the population within that age range. Table 1 shows the statistics of the patient demographics. Unlike the older CRC patient population between age 50 and 64,[2] more women (66.34%) were found in our CRC patient group, and 6% of them were pregnant. Their regular prenatal checkups provided more than 7 months of patient journey in EHR and consequently made them eligible in our study.

Table 1. Statistics of CRC patient demographics

| Age | Mean (SD) 33.35 (6.88) |
|---|---|
| Gender | Male 33.66%. Female 66.34% |
| Race | White: 61.51%, Black: 16.51%, Asian: 5.39%, Other: 9.56%, Not Specified: 7.02% |
| Ethnicity | Not Hispanic or Latino: 80.35%, Hispanic or Latino: 18.18%, Mexican or Puerto Rican: 1.47% |

We created a balanced training set with 1853 CRC patients and 1853 Non-CRC patients randomly sampled from the population. The idea was to improve the predictive performance on the CRC group by under-sampling the overpresented Non-CRC patients. Some studies detecting rare diseases applied synthetic minority over-sampling technique (SMOTE) to up-sample the rare cases by interpolation in feature space.[20,21] Since LLMs process natural languages rather than numeric feature vectors, synthetic longitudinal patient events with realistic EHR medical coding would be needed to provide augmented samples for the CRC patient group.[22] We leave this for future work. For the test set, we split the rest of the 100 CRC patients into 10 different runs, each combined with 990 Non-CRC patients, such that each test run comprised 1% of CRC patients. We reported the average sensitivity (or recall), specificity, precision (or positive predictive value (PPV)), negative predictive value (NPV) and F1 measures across these 10 test runs with 95% confidence intervals to assess the general model performance.

Statistical Machine Learning models: For the ML approaches, various classification algorithms were employed, including logistic regression (LR), k-Nearest Neighbors (KNN), naive bayes (NB), support vector classifier (SVC), decision tree (DT), random forest (RF), adaptive boosting (AdaBoost), light gradient boosting machine (LightGBM), histogram-based gradient boosting (HGBoost), and extreme gradient boosting (XGBoost). Each model was optimized using random search with 5-fold cross-validation on the training set, selecting hyperparameters based on the highest F1-score. As shown in Table 2, gradient boosting methods (AdaBoost, LightGBM, HGBoost, and XGBoost) and LR achieved the highest F1-scores (≥0.824), outperforming other classifiers.

Table 2. Performance of 5-fold cross-validation on the training set

| Model | LR | KNN | NB | SVC | DT | RF | AdaBoost | LightGBM | HGBoost | XGBoost |
|---|---|---|---|---|---|---|---|---|---|---|
| F1-Score | 0.824 | 0.607 | 0.661 | 0.740 | 0.725 | 0.787 | 0.842 | 0.832 | 0.855 | 0.838 |

All available features, including conditions, laboratory results, and observational data from EHR, were incorporated to examine the intricate relationships between variables comprehensively. Missing values were not removed; instead, they were treated as "not specified" to preserve potential signal. Feature engineering involved transformation of demographic variables into one-hot encoded vectors, indicating their presence or absence. Conditions were represented by how often they were diagnosed within the relevant time period. For continuous lab results and condition values, mean values were used.

To prevent information leakage, the training and test sets were processed independently, leading to variations in feature counts across datasets. The training set contained 11,509 features, while the 10 test sets had between 5,209 and 5,783 features. This discrepancy stemmed from differences in sample sizes. To ensure a consistent feature space across all models, any features present in the training set but missing in a particular test set were added to that test set, with values set to zero. To mitigate the potential for overfitting due to this sparsity, feature regularization techniques were applied during model training (e.g., L1 or L2 regularization, depending on the specific algorithm).[23]

The decision threshold for each model was optimized to balance sensitivity (recall) and specificity, given the very low prevalence of CRC cases in our test sets. The optimal threshold was determined using Youden's J Index (J = Sensitivity + Specificity - 1) calculated from the Receiver Operating Characteristic (ROC) curve.[24,25] To ensure a robust and generalizable threshold, and to avoid overfitting, we performed 10-fold cross-validation, using imbalanced validation sets within each fold that mimicked the target CRC proportion (approximately 1% CRC cases).

Large Language Models: For the LLM approach, we employed OpenAI's GPT-4o (version released 2024-08-06). Patient information, including demographics, was input as plain text. Medical conditions, lab results and observations were represented by the names of their corresponding ICD, SNOMED-CT and LOINC codes in order to capture the clinical context. Although patients could be reported with the same conditions or observations from multiple encounters within 6 months prior to the onset of CRC, only one instance of conditions and observations would be used in the input when prompting the LLM. The values of the lab results and their units of measure (when available) were included. However, only the latest value of the same lab tests would be used.

The prompt comprised five components: 1) definition of the AI assistant's role; 2) Chain-of-Thought[26] reasoning steps for CRC determination; 3) general guidelines for CRC diagnosis, synthesized from the U.S. Preventive Services Task Force recommendations;[27] 4) a specified output format; and 5) patient-specific input. This template, shown in Figure 1, was designed to facilitate the processing of complex, long-context patient information that required extended reasoning and holistic consideration. Wong et al. proposed similar prompting approach, namely UNIMEDABSTRACTOR (UMA), to extract information from clinical notes.[28] When prompting LLMs, we set max tokens to 4096 and temperature to 0 in the chat completion, in addition to all the default parameters.

We evaluated the performance of the base GPT-4o LLM and also the fine-tuned version. Using the same training set as in the ML approaches, we fine-tuned the LLM for 2 epochs.

```
You are a healthcare clinical expert who can identify and stage colorectal cancer. Your task is to predict the possibility of
colorectal cancer developing in the future, given the patient conditions and lab results. and observations. The patient
conditions are given in this format: CONDITIONS: [cond1, cond2, etc]. The lab results are given in this format: LAB RESULTS:
[lab1 (value), lab2 (value), etc]. The observations are given in this format: OBSERVATIONS: [obs1 (value), obs2 (value), etc].
When making your prediction, use step-by-step reasoning. Carefully evaluate the patient's conditions, lab results and
observations against the relevant factors outlined in the colorectal cancer guidelines. Break down your reasoning into the
following steps:
   1. Identify all relevant patient's conditions, lab results and observations.
   2. Compare these findings to the symptoms, signs, condition, lab tests and risk factors provided in the guidelines.
   3. Consider the severity of conditions, lab results values (in parenthesis) and observations values (in parenthesis) to
determine the likelihood of colorectal cancer.
   4. Include only the following in your response:

   - Based on your reasoning, answer only yes if there is high risk that the patient may develop colorectal cancer in the next
6 months, otherwise answer no. Give the answer in this format: Answer: X
   - Next, provide a probability score (from 0% to 100%) indicating how likely the patient is to develop colorectal cancer, in
this format: Probability score: X%
   - Finally, explain your answer using the provided patient conditions, lab results, observations and the guidelines to
support your reasoning.
   - Do not use markdown or templates that include **

   <Guidelines>

   1. Symptoms and Signs
   - Early Stages:
   * Changes in bowel habits (diarrhea, constipation, narrow stool)
   * Persistent feeling of needing a bowel movement
   * Rectal bleeding (bright red) or blood in stool (dark brown/black)
   * Cramping or abdominal pain
   * Weakness, fatigue, unintended weight loss
   * Anemia (low red blood cell count)
   - Advanced Stages:
   * Enlarged liver
   * Jaundice (yellowing of the skin/eyes)
   * Difficulty breathing (due to cancer spreading to lungs)

   2. High-Risk Conditions
   * Malignant neoplastic disease
   * Fresh blood passed per rectum
   * Hemorrhage of rectum and anus
   * Colitis
   * Pain due to neoplastic disease
   * Hemorrhoids
   * Melena
   * Malignant tumor of anus
   * Rectal hemorrhage
   * Localized enlarged lymph nodes
   * Altered bowel function
   * Human immunodeficiency virus infection
   * Gastrointestinal hemorrhage
   * Intra-abdominal and pelvic swelling, mass and lump
   * Rectal pain
   * History of disorder of digestive system
   * Chronic constipation

   3. High-Risk Lab tests
   * Glomerular filtration rate/1.73 sq M.predicted Volume Rate/Area
   * Carbon dioxide, total Moles/volume
   * Cholesterol in HDL Mass/volume
   * Erythrocyte distribution width Ratio
   * Chloride Moles/volume
   * Monocytes/100 leukocytes
   * Potassium Moles/volume
   * Alkaline phosphatase Enzymatic activity/volume
   * Prothrombin time (PT)
   * Anion gap

   4. High-Risk observations
   * Significant weight change
   * Pain severity - 0-10 verbal numeric rating Score - Reported
```

Figure 1. Prompt template to predict onset of CRC

Results

Table 3 shows a comparative analysis of the average performance across 10 test runs for both the ML models and the LLMs. LLMs generally achieved better performance than statistical ML models, though all models expectedly exhibited limited precision and F1 score given the dataset's class imbalance (since CRC is a rare disease).

Among the ML approaches, LR, XGBoost, and SVC provided relatively balanced performance between sensitivity and specificity. RF also exhibited a balanced performance that was comparable to that of GPT4-o (Base). In contrast, methods such as LightGBM and HGBoost tended to achieve very high sensitivity at the expense of specificity, indicating potential overfitting to positives and making them less reliable for detecting rare events. Gradient boosting methods, while often excelling in other applications, are known to struggle with extreme class imbalance, such as the 1% CRC prevalence observed in this dataset.

Furthermore, supervised fine-tuning allowed LLM to learn the specific patterns in the training data and enhance overall performance.[29] As can be observed, the fine-tuned GPT4-o model improved sensitivity or recall by more than 10%. On the other hand, we also observed relatively low precision and F1 score in all models. Similar observations were also reported by Zhen et al. in their EoCRC study in China.[10] This was mainly due to the highly imbalanced nature of the data and the low prevalence (1%) of CRC in the test set. Nonetheless, the fine-tuned LLM achieved 73% sensitivity with 91% specificity and 99.7% negative predictive value. This indicated that the model could potentially be used to exclude those patients who were identified by the model as low risk from cancer screening, minimizing unnecessary cost, while over 70% of the CRC patients could correctly be detected by the model. It is worth noting that even though we attempted to prompt GPT4-o to provide a probability score on the prediction, the generated scores were presented as a numerical form of Likert Scale, i.e. 30%, 50%, 75%, etc. Hence, we could not consider the scores in any statistical context, and use them to plot the ROC curve.

In summary, while the ML models struggled to balance sensitivity and specificity due to the extreme class imbalance, the fine-tuned GPT4-o offered the most reliable trade-off, suggesting its potential utility in real-world screening scenarios.

Table 3. Average performance of machine learning and large language models across 10 test runs

| Model | Sensitivity (Recall) | Specificity | Precision (PPV) | NPV | F1-Score |
| --- | --- | --- | --- | --- | --- |
| LR | 0.910 ±(0.046) | 0.642 ±(0.007) | 0.025 ±(0.001) | 0.998 ±(0.001) | 0.049 ±(0.003) |
| KNN | 0.440 ±(0.073) | 0.864 ±(0.005) | 0.032 ±(0.004) | 0.994 ±(0.001) | 0.059 ±(0.008) |
| NB | 0.900 ±(0.046) | 0.355 ±(0.009) | 0.014 ±(0.001) | 0.997 ±(0.001) | 0.028 ±(0.001) |
| SVC | 0.810 ±(0.068) | 0.673 ±(0.006) | 0.024 ±(0.002) | 0.997 ±(0.001) | 0.047 ±(0.004) |
| DT | 0.520 ±(0.064) | 0.785 ±(0.008) | 0.024 ±(0.003) | 0.994 ±(0.001) | 0.046 ±(0.005) |
| RF | 0.650 ±(0.067) | 0.814 ±(0.038) | 0.037 ±(0.008) | 0.996 ±(0.001) | 0.070 ±(0.014) |
| AdaBoost | 0.740 ±(0.084) | 0.333 ±(0.035) | 0.011 ±(0.002) | 0.992 ±(0.003) | 0.022 ±(0.004) |
| LightGBM | 1.000 ±(0.000) | 0.084 ±(0.082) | 0.011 ±(0.001) | 0.700 ±(0.299) | 0.022 ±(0.002) |
| HGBoost | 1.000 ±(0.000) | 0.005 ±(0.006) | 0.010 ±(0.000) | 0.400 ±(0.320) | 0.020 ±(0.000) |
| XGBoost | 0.910 ±(0.068) | 0.559 ±(0.121) | 0.025 ±(0.008) | 0.998 ±(0.001) | 0.049 ±(0.015) |
| GPT-4o (Base) | 0.620 ±(0.121) | 0.893 ±(0.008) | 0.055 ±(0.011) | 0.996 ±(0.001) | 0.102 ±(0.021) |
| GPT-4o (FT) | 0.730 ±(0.096) | 0.910 ±(0.007) | 0.076 ±(0.009) | 0.997 ±(0.001) | 0.138 ±(0.017) |

Explainability

Despite the promising predictive capability of ML models, many argue that explainability of models is an indispensable quality to justify their performance. From the medical perspective, Amann et al. suggested two levels of explainability, specifically how the models reached the overall conclusive results, and which features were important to contribute to the prediction of each individual sample.[30] We investigated these two levels of explainability in one of the best performing statistical ML models, XGBoost. Figure 2 lists the importance of the top 20 features used by the model to predict sample outcomes, based on the average gain when the feature was split at the tree node. In particular, "carcinoembryonic lab test", "phosphate lab test", "malignant neoplastic disease condition", "oxygen saturation lab test" and "fresh blood passed per rectum" are the top 5 informative features.

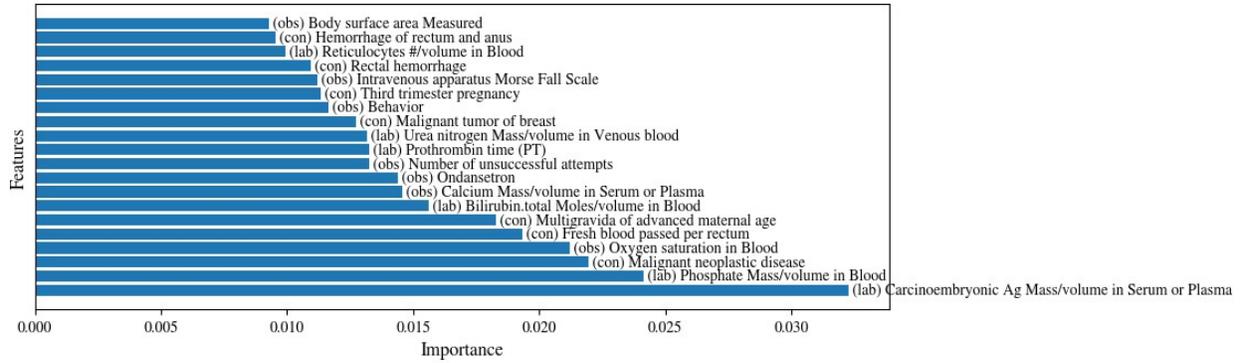

Figure 2. Feature importance of the XGBoost model

In addition, SHAP (SHapley Additive exPlanations) values[31] were calculated to provide instance-level feature importance explanation. Figure 3 shows the waterfall plot of the most impactful features and how each feature influenced the prediction of the outcome for one CRC patient. More specifically, "Erythrocyte distribution width Ratio" increased most of the model's predicted probability of having CRC outcome, while "Glomercular filtration rate" reduced the predicted probability.

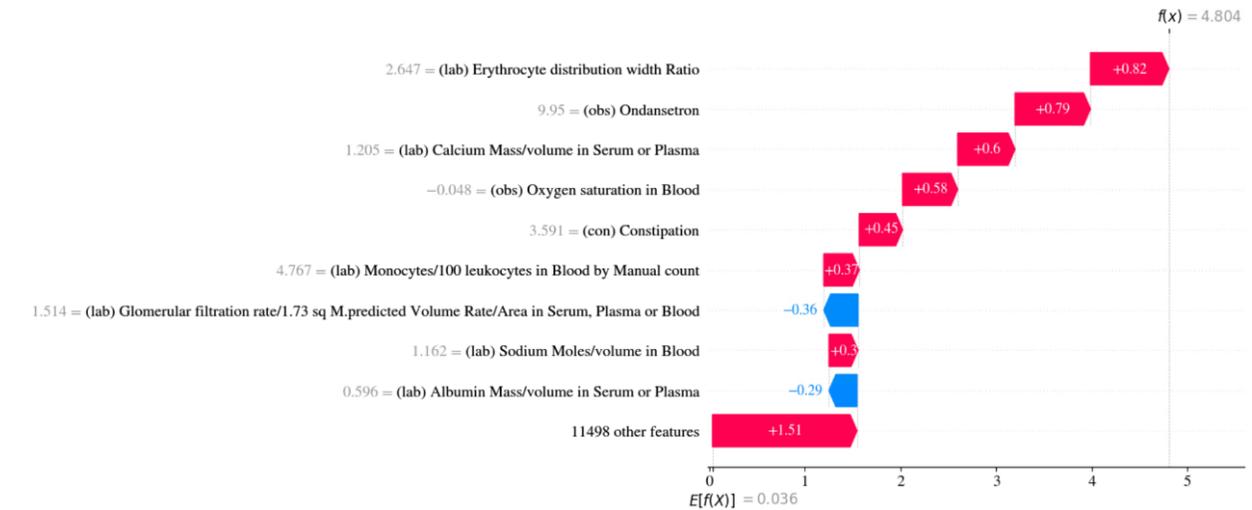

Figure 3. Contribution of top features to the model prediction on a CRC patient

While feature importance and SHAP values offered analytical interpretation of how a model arrived at a prediction, LLM on other hand could provide a more intuitive explanation in the generated text. Such model behavior was referred to as Natural Language Explanation by Zhao et al.[32] One study showed that Chain-of-Thought prompting could lead to more stable reasoning and consistent answers.[33] Figure 4 shows the explanation from the LLM on predicting the outcome of a CRC patient based on the prompt provided in Figure 1.

```
1. Identify all relevant patient's conditions, lab results, and observations.

Conditions:
- Constipation
- Abdominal pain
- Severe protein-calorie malnutrition
- History of antineoplastic chemotherapy
- H/O: chemotherapy
- Body mass index less than 20

Lab Results:
- Glomerular filtration rate/1.73 sq M.predicted (128.00)
- Chloride (109.00)
- Alkaline phosphatase (97.00)
- Carbon dioxide, total (19.00)
- Anion gap (9.00)
- Monocytes/100 leukocytes (8.40)
- Potassium (4.30)

Observations:
- Body mass index (BMI) [Ratio] (17.90)
- Pain intensity (9.00)

2. Compare these findings to the symptoms, signs, condition, lab tests, and risk factors provided in the guidelines.

Symptoms and Signs:
- The patient has abdominal pain and constipation, which are early-stage symptoms of colorectal cancer.
- Severe protein-calorie malnutrition and a BMI less than 20 indicate significant weight change, a high-risk observation.

High-Risk Conditions:
- Constipation is a high-risk condition.
- History of antineoplastic chemotherapy and H/O: chemotherapy suggest a history of cancer treatment, which may increase the risk of cancer recurrence.

High-Risk Lab Tests:
- Glomerular filtration rate and Chloride are high-risk lab tests.
- Alkaline phosphatase is also a high-risk lab test.

High-Risk Observations:
- Pain intensity of 9.00 is a high-risk observation indicating severe pain.

3. Consider the severity of conditions, lab results values, and observations values to determine the likelihood of colorectal cancer.

- The presence of constipation and abdominal pain, along with a history of chemotherapy, suggests a potential risk for colorectal cancer.
- Severe protein-calorie malnutrition and a low BMI indicate significant weight change, which is a high-risk observation.
- High-risk lab tests such as Glomerular filtration rate, Chloride, and Alkaline phosphatase further increase the risk.
- Severe pain intensity (9.00) is concerning and aligns with high-risk observations.

Based on the above analysis, the patient exhibits multiple high-risk factors and symptoms associated with colorectal cancer.

Answer: Yes
Probability score: 75%
The patient has several high-risk conditions, lab results, and observations that align with the guidelines for colorectal cancer risk. The combination of constipation, abdominal pain, history of chemotherapy, severe malnutrition, and high-risk lab results and observations suggests a significant likelihood of developing colorectal cancer in the next 6 months.
```

Figure 4. Explanation from LLM on predicting the outcome of a CRC patient

Conclusion

Given the rising incidences of CRC in younger individuals, the American Cancer Sociey (ACS) lowered the age recommendation for stool-based test and structural examination from 50 to 45.[34] One study based on Markov model analysis suggested that initiating CRC screening at age 45 years could be cost effective, and could prevent 29,400 CRC cases and 11,100 CRC deaths over the next 5 years, despite the additional cost of $10.4 billion in colonoscopies.[35] Discussion on the implementation of this recommendation at the national level remained inconclusive. In particular, the screen pool would be anticipated to increase by 22%, and the balance of screening burden to benefit was unclear.[36,37] This same argument could extend and be manifested to an even younger age group (18-44), with a larger

burden-to-benefit ratio. This presents an opportunity to leverage machine learning to alleviate the burden of screening cost for the younger population.

In this context, our study demonstrated the potential of ML and LLMs, specifically a fine-tuned GPT-4o, to predict early-onset CRC (EoCRC) risk using electronic health record (EHR) data. Statistical ML models (LR, XGBoost, and SVC) detected 81–91% of CRC cases but correctly identified only 56–64% of non-CRC patients, resulting in lower F1 scores and precision compared to LLMs. Interestingly, the base GPT-4o model performed comparably to RF, despite receiving no explicit training data. This highlights the inherent medical knowledge embedded within LLMs and their potential applicability in clinical settings. While the GPT-4o base model contained general medical knowledge, supervised fine-tuning with explicit guideline enhanced its predictive capabilities on EoCRC. The fine-tuned LLM was able to detect on average 73% cases of CRC patients, and 91% of Non-CRC patients. The accuracy of identifying the Non-CRC cases was at 99.7%. The performance was comparable to faecal immunochemical tests (FITs), while the sensitivity was 34% higher than the results from guaiac fecal occult blood tests (gFOBTs) for colorectal cancer screening (age ≥ 40).[38] Furthermore, the explanation in natural language from the LLM was easier to understand than the degree of feature contribution in statistical ML models.

Despite these promising findings, this study has some limitations. First, it is a retrospective study where the patient information was extracted from the coded data in EHR. The inherent information bias could be due to missing data issues and inaccurate coding.[39] Second, we only experimented with OpenAI's flagship model, GPT4-o LLM. In future work, we plan to evaluate other LLMs, like Meta's Llama3 and Google's Gemini to gain better understanding of LLMs' reasoning capability in CRC prediction. Lastly, the study was limited to the patients with 7 months of medical history of clinical data including conditions, lab results and observations, in order to provide sufficient training signals for the models to learn. However, we found many cases where the patients were diagnosed CRC within one month after their first visit and therefore they were not eligible to be in the study. This could be attributed to the general good health in the younger population, and the absence of other health issues that required medical attention. Many patients also had irregular visits, or no visits in a few months. The data sparsity in patient journey with irregular patterns could be challenging to learn. One potential avenue for future research is to pre-train a clinical language model that captures the nuances in different patient journeys through a large amount of longitudinal medical records. By learning the joint probability distribution between events in the patient journeys, better performance could be achieved without abundant and highly accurate training data.[40,41]

Acknowledgments

We thank Michael Simonov, Sara Daraei, Rajesh Rao, and Jay Pillai for helpful discussions. This work is supported by Truveta.